\documentclass[10pt,twocolumn,letterpaper]{article}

\usepackage[pagenumbers]{cvpr} 

\usepackage[table]{xcolor}
\usepackage{makecell}
\usepackage{arydshln}
\usepackage{multirow}
\usepackage{graphicx}
\usepackage{capt-of} 
\definecolor{colorfirst}{RGB}{255,153,153}
\definecolor{colorsecond}{RGB}{255,204,153}
\definecolor{colorthird}{RGB}{255,255,153}

\usepackage{fontawesome5}

\makeatletter
\renewcommand\paragraph{\@startsection{paragraph}{4}{\z@}{0.7ex plus 0.2ex minus 0.1ex}{-1em}{\normalfont\normalsize\bfseries}}
\makeatother

  {\par} 

\usepackage{enumitem}
\setlist{nosep} 

\definecolor{cvprblue}{rgb}{0.21,0.49,0.74}
\usepackage[pagebackref,breaklinks,colorlinks,citecolor=cvprblue]{hyperref}

\def\paperID{****} 
\def\confName{CVPR}
\def\confYear{****}

\title{BulletTime: Decoupled Control of Time and Camera Pose for Video Generation}

\author{
Yiming Wang$^{1,2}$ \quad
Qihang Zhang$^{3,2*}$ \quad
Shengqu Cai$^{2*}$ \quad
Tong Wu$^{2\dagger}$ \quad
Jan Ackermann$^{2\dagger}$ \\
Zhengfei Kuang$^{2\dagger}$ \quad
Yang Zheng$^{2\dagger}$ \quad
Frano Raji\v{c}$^{1\dagger}$ \quad
Siyu Tang$^{1}$ \quad
Gordon Wetzstein$^{2}$ \\
\\[0.1em]
$^{1}$ETH Zurich \quad
$^{2}$Stanford University \quad
$^{3}$CUHK \\
\\[0.1ex]
{\fontsize{11.00pt}{11.5pt}\selectfont
{\href{https://19reborn.github.io/Bullet4D/}{\texttt{https://19reborn.github.io/Bullet4D/}}}
}
}

\begin{document}

\twocolumn[{%
  \renewcommand\twocolumn[1][]{#1}
  \maketitle
  \begin{center}
    \vspace{0.4cm}
    \captionsetup{type=figure}
    \includegraphics[width=\textwidth]{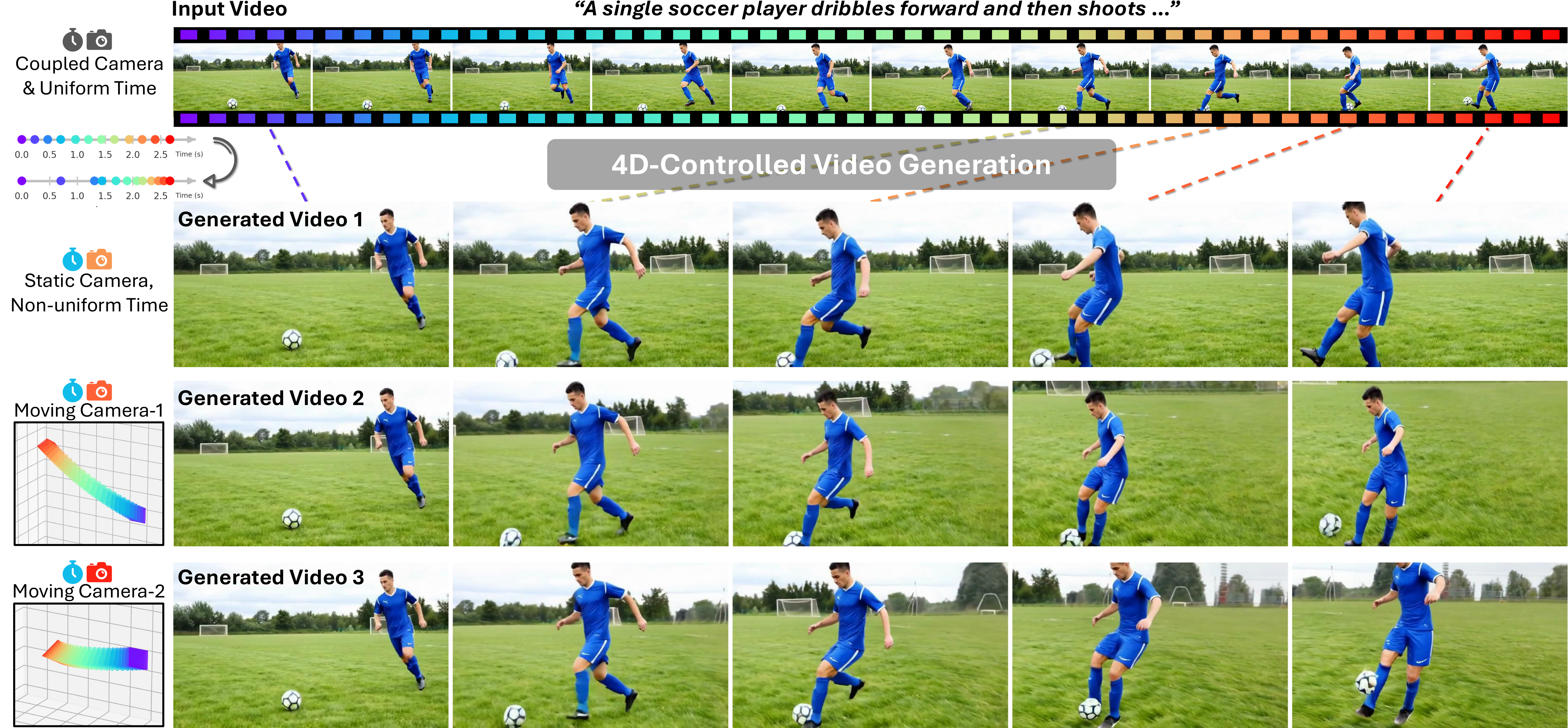}
    \captionof{figure}{\textbf{Time- and camera-controlled 4D video generation.}
    Given a single input video where camera motion is entangled with uniform temporal sampling (top row), our method synthesizes new videos that enable decoupled control over world time and camera pose. }
    \label{fig:teaser}
  \end{center}
}]

\let\svthefootnote\thefootnote
\newcommand\freefootnote[1]{%
  \let\thefootnote\relax%
  \footnotetext{#1}%
  \let\thefootnote\svthefootnote%
}

\freefootnote{$*$, $\dagger$ Equal contribution}

\begin{abstract}
Emerging video diffusion models achieve high visual fidelity but fundamentally couple scene dynamics with camera motion, limiting their ability to provide precise spatial and temporal control. We introduce a 4D-controllable video diffusion framework that explicitly decouples scene dynamics from camera pose, enabling fine-grained manipulation of both scene dynamics and camera viewpoint. Our framework takes continuous world-time sequences and camera trajectories as conditioning inputs, injecting them into the video diffusion model through a 4D positional encoding in the attention layer and adaptive normalizations for feature modulation. To train this model, we curate a unique dataset in which temporal and camera variations are independently parameterized; this dataset will be made public. Experiments show that our model achieves robust real-world 4D control across diverse timing patterns and camera trajectories, while preserving high generation quality and outperforming prior work in controllability. 
\end{abstract}    
\section{Introduction}
\label{sec:intro}
Recent advances in diffusion models have enabled photorealistic video generation~\cite{gupta2024photorealistic, Sora, Veo, pika, Kling}. The temporal dynamics in these models are typically represented in \textit{video time}, derived from frame indices and frame rate. This representation conflates two fundamentally different dimensions: \textit{world time}, denoting the absolute temporal coordinate that governs the evolution of the scene dynamics, and \textit{camera pose}, determining the viewpoint from which the scene is observed.
Decoupling these quantities would allow users to independently control \textit{when} and \textit{where} to observe events within a dynamic 4D world, enabling applications ranging from cinematic effects (e.g., bullet time) to gaming and XR scenarios where users freely navigate a frozen or slowed-down scene.
Domains including free-viewpoint video synthesis~\cite{van2024gcd, yu2025trajectorycrafter, bai2025recammaster} and robotics~\cite{liu2025geometry, li2025unified,wen2024vidman, rigter2024avid} would also benefit from the aforementioned capabilities. 
Therefore, our goal is to equip video diffusion models with disentangled control over these two dimensions, paving the way toward 4D world modeling~\cite{genie3,feng2024matrix, WorldLabs} and simulation.
Recent camera-controlled video diffusion models~\cite{wang2024motionctrl,he2024cameractrl,he2025cameractrl2,bai2025recammaster,yu2025trajectorycrafter} provide a step towards our vision by explicitly conditioning on camera trajectories. While enabling viewpoint control, these methods still entangle video time with world time by treating the frame index as the implicit physical time of the scene. This assumption enforces a uniformly advancing world time tied to frame indices, preventing independent manipulation of continuous temporal effects such as slowing down or pausing scene dynamics independently of camera movement. 
Moreover, naively extending text prompts to control time cannot provide fine-grained or accurate control over world time, and a two-stage pipeline that first remaps the input video and then applies camera-controlled video diffusion often results in poor 4D consistency (see Sec.~\ref{sec:experiments}).
Multi-view video diffusion methods~\cite{wu2025cat4d} further rely on a separate reconstruction step built from extensively sampled generated videos (e.g., 4D Gaussian Splatting~\cite{wu20244dgs}) to achieve full camera- and time control, making interactive 4D world modeling infeasible.
To address these limitations, we introduce a 4D-controllable video diffusion model that decomposes changes over video time into continuous world time and camera viewpoint, used as explicit conditioning signals for the generation process. 
To effectively condition on continuous world time, we introduce two complementary mechanisms: a time-aware positional encoding \textbf{(Time-RoPE)} that injects world-time information into the attention mechanism, and a time-conditioned adaptive layer normalization \textbf{(Time-AdaLN)} module that provides fine-grained temporal modulation. 
We conduct ablation studies to demonstrate the effectiveness of our design compared to alternative mechanisms for world-time conditioning.
To extend temporal control to full 4D disentanglement, we incorporate a camera-conditioned adaptive normalization layer together with a unified 4D positional encoding (\textbf{4D-RoPE}), augmenting the camera-aware rotary transformations in Time-RoPE.
To support disentangled learning and evaluation, we curate a 4D-controlled synthetic dataset in which temporal and camera factors vary independently, enabling robust generalization across diverse trajectories, timing patterns, and real-world scenarios. 
%

%
In summary, the main contributions of our paper are as follows:

 \begin{itemize}

\item We introduce a 4D-controllable video diffusion framework that disentangles visual evolution along video time into world time and camera pose, and conditions the model using a unified 4D positional encoding together with adaptive normalization modules.

\item We curate a 4D-controlled synthetic dataset with independently varied temporal and camera factors to support both training and evaluation, facilitating the learning of disentangled camera–time control.

\item Ablations demonstrate the superiority of our conditioning design over alternative strategies. Our method generalizes to diverse real-world scenarios and 4D control inputs, achieving state-of-the-art performance in 4D-controllable video-to-video generation.

\end{itemize}

\section{Related Work}

\subsection{Video Diffusion Models}
\looseness=-1
Diffusion models have achieved remarkable success in image synthesis~\cite{ho2020denoising, song2019generative, song2020denoising, rombach2022high, peebles2023scalable, dhariwal2021diffusionbeatgan, ramesh2022hierarchical} and have recently been extended to video generation~\cite{ho2022video,he2022latentvideodiffusion,singer2022makeavideo, guo2023animatediff, blattmann2023svd, ho2022imagenvideo, blattmann2023align, Sora, Veo, yang2024cogvideox, wan2025}, enabling realistic synthesis from diverse conditional inputs.
Early video diffusion models~\cite{ho2022video, he2022latentvideodiffusion, singer2022makeavideo, blattmann2023align, blattmann2023svd, ho2022imagenvideo, guo2023animatediff} extended U-Net--based image generators~\cite{ronneberger2015unet} with temporal modules across frames, while recent approaches~\cite{yang2024cogvideox, kong2024hunyuanvideo, wan2025} adopt transformer architectures~\cite{esser2024scaling, peebles2023scalable} with 3D full attention for improved spatiotemporal coherence and visual quality.

\subsection{Camera-Controlled Video Generation}
\looseness=-1
Beyond conditioning on text prompts or images, camera-controlled video diffusion methods leverage camera poses to enable global viewpoint manipulation over general scene content, offering a natural interface for scene exploration~\cite{he2025cameractrl2} and world modeling~\cite{wu2025videoworld,genie3}.
Multi-view video diffusion models further extend this capability by synthesizing synchronized videos from multiple camera views~\cite{kuang2024cvd,xu2024cavia,li2024vivid,bai2024syncammaster,wu2025cat4d}.

\subsection{Time-Controlled Video Generation}
\looseness=-1
Most video diffusion models encode the temporal dimension by applying positional encodings to discrete frame indices, and offer limited support for continuous and precise control of the underlying physical world time.
Commercial systems such as Stable Video Diffusion~\cite{blattmann2023svd} and Wan~\cite{wan2025} provide only coarse motion-strength adjustments, modulating the amount of motion but not exposing an explicit, controllable temporal signal.
Motion-controlled video diffusion methods guide scene dynamics using motion trajectories~\cite{burgert2025gowiththeflow,chang2024how,yang2024direct,wang2024motionctrl,shi2024motion,wu2024draganything,namekata2024sg,qiu2024freetraj,geng2024motion,yin2023dragnuwa,xiao2024trajectoryattn,zheng2025vidcraft3}, pose skeletons~\cite{ma2024follow,ruiz2023dreambooth,chang2025xdyna,guo2023animatediff,hu2024animate}, or reference videos~\cite{wang2024videocomposer,yatim2024space,geyer2023tokenflow,yin2024scalable,ling2024motionclone,mou2024revideo}. 
While these approaches control object motion for applications such as video editing, they do not render a single coherent scene across different world times.
Cat4D~\cite{wu2025cat4d} and 4DiM~\cite{watson20244dim} propose unified 4D formulations that jointly model camera and time within the diffusion framework, enabling training on mixed static and dynamic data. Yet 4DiM, being image-based, cannot produce high-quality 4D-controllable videos.
Cat4D and many other multi-view video diffusion models~\cite{zhang2024fourdiffusion, lin2024alignyourgaussians, sun2025dimensionx, wu2025cat4d} employ an additional 4D reconstruction stage to render scenes across camera poses and times; however, this process typically necessitates extensive multi-view sampling.
A concurrent work, Lyra~\cite{bahmani2025lyra}, distills video diffusion models to decode explicit 3D/4D Gaussian scene representations. 
However, these approaches inherit the uniformly discretized temporal of the underlying video diffusion model, limiting its ability to interpolate high-quality frames at novel times.
Complementary to these directions, we introduce a time-aware positional encoding that overcomes this limitation by enabling continuous temporal control in video diffusion models. We further curate a 4D synthetic dataset, enabling real-world generalization with explicit and independent control of camera motion and world time.
Despite being fine-tuned solely on synthetic data, our method generalizes effectively to real-world scenarios, and produces 4D-controlled videos without requiring explicit post-processing or auxiliary 4D representations.


\begin{figure*}[t]
  \centering
  \includegraphics[width=1.0\linewidth]{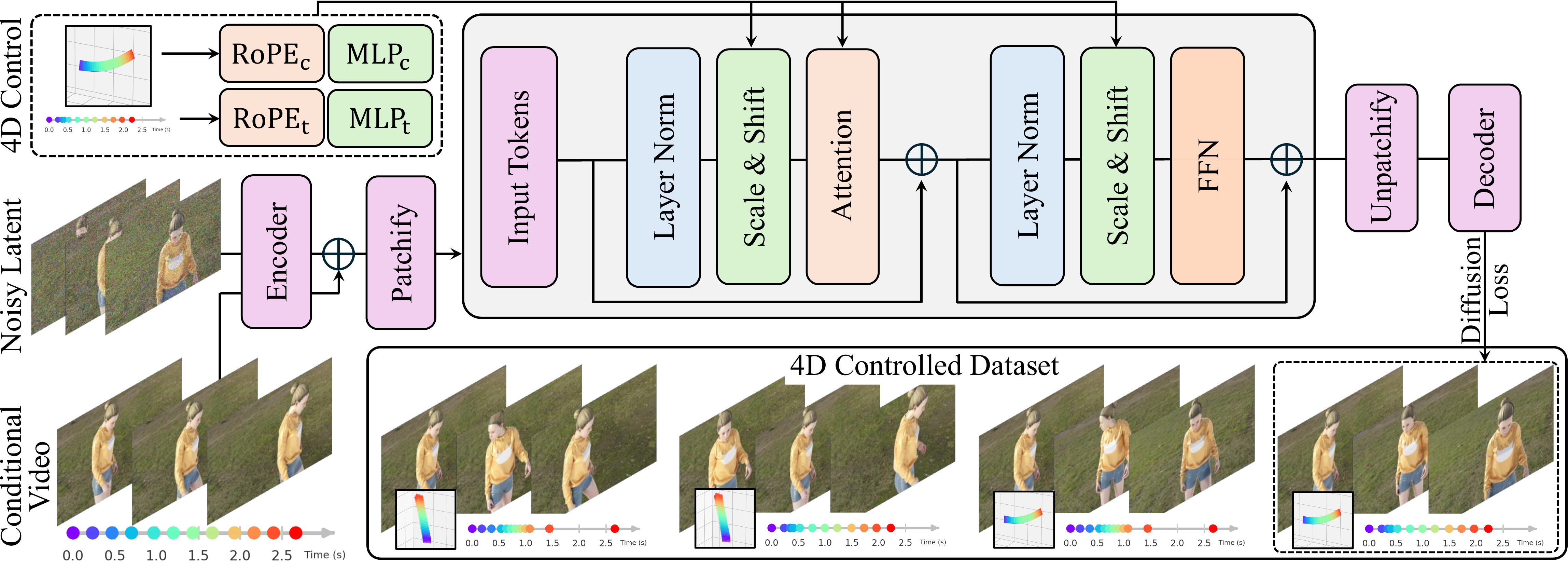} 
  \vspace{-0.7cm}
\caption{
    \textbf{Method Overview.} Given a conditional input video, our diffusion model generates new videos under 4D control using world time and camera trajectory. These two signals are injected into the Diffusion Transformer through complementary modulation pathways. Time control is enabled by $\mathrm{RoPE}_t$ (a time-aware positional encoding injected into attention) and $\mathrm{MLP}_t$, which predicts the affine scale and shift used to modulate intermediate features. Camera control is introduced analogously through $\mathrm{RoPE}_c$ (a camera-aware positional encoding) and $\mathrm{MLP}_c$. The outputs of $\mathrm{RoPE}_t$ and $\mathrm{RoPE}_c$ are fused into a unified 4D positional encoding injected into the attention layers. Together, these mechanisms form a 4D-controllable DiT block capable of jointly steering temporal evolution and camera motion during generation. We train our model on a curated 4D-controlled synthetic dataset that we constructed, where temporal and camera factors vary independently across scenes, providing explicit supervision for disentangling time and camera control.
}
  \label{fig:overview}
\end{figure*}

\section{Method}
\label{sec:method}

Our goal is to enable \emph{4D-controllable video generation}, where world time and camera viewpoint can be explicitly manipulated during synthesis. Conventional video diffusion models couple these two dimensions: the apparent motion in generated videos jointly reflects scene dynamics and camera movement along a shared video-time axis. We reformulate this by introducing two explicit and orthogonal conditioning signals, \textbf{world time} $\tau_{\text{world}}$ and \textbf{camera pose} $c$, which together define a controllable 4D coordinate system.

Our framework, illustrated in \cref{fig:overview}, consists of three components: a continuous world-time control mechanism (\cref{subsec:method:time_control}), a unified 4D time–camera conditioning module (\cref{subsec:method:camera_control}), and a 4D-controlled dataset that provides supervision for learning disentangled control (\cref{subsec:method:dataset}).

\subsection{Preliminaries: Video Diffusion Models}
\label{subsec:method:preliminary}
A video diffusion model transforms random noise into coherent video sequences through a learned denoising process, typically operating in a compact latent space encoded by a pretrained 3D VAE. Diffusion Transformers (DiTs)~\cite{peebles2023scalable} have become the standard denoiser architecture in recent video diffusion models. DiTs divide the latent video tensor $\tilde{z}_{t}$ into non-overlapping spatiotemporal patches and processes them using attention~\cite{vaswani2017attention}:
\begin{equation}
\mathrm{Attn}(Q, K, V) = \mathrm{softmax}\left(\tfrac{QK^\top}{\sqrt{d}}\right)V,
\label{eqn:attention}
\end{equation}
where $Q,K,V\!\in\!\mathbb{R}^{N\times d}$ are the query, key, and value matrices, $N$ denotes the number of spatiotemporal tokens (patches), and $d$ is the feature dimension of each token. Because attention is permutation-invariant, positional encodings are required to provide ordering information and are commonly implemented as rotary positional embeddings (RoPE)~\cite{shaw2018relPE,su2024roformer}. Recent video diffusion models~\cite{kong2024hunyuanvideo,yang2024cogvideox} typically adopt a 3D extension of RoPE, where independent rotations are applied along time, height, and width.

\subsection{Continuous-Time Control}
\label{subsec:method:time_control}
We aim to enable explicit temporal control in video diffusion models, allowing users to specify the exact physical time at which each frame should be generated. To achieve this, we condition the model on a continuous time sequence $\{\tau_i\}_{i=0}^{F-1}$, where each $\tau_i \in \mathbb{R}^+$ denotes the world time associated with frame $i$ and $F$ is the number of frames. Varying the temporal interval $\tau_{i+1}-\tau_i$ enables arbitrary time reparameterizations such as slow motion, acceleration, temporal pausing, or reversal. We incorporate the continuous-time signal into the Diffusion Transformer (DiT) through (i) a time-aware positional encoding (Time-RoPE) and (ii) a time-conditioned adaptive normalization module (Time-AdaLN).

\paragraph{Time-RoPE.}
Conventional video diffusion models represent temporal evolution using positional encodings tied to \emph{discrete} frame indices. This discretization inherently assumes uniform temporal spacing and prevents the model from responding to fine-grained control. To resolve this limitation, we extend RoPE~\cite{su2024roformer} to operate directly on \textit{continuous} time and refer to this extension as Time-RoPE.

Formally, we define the time-aware rotation operator $\mathbf{D}^{\text{Time}}(\tau)\!\in\!\mathbb{R}^{d'\times d'}$ as the following block-diagonal matrix:
\begin{equation}
\label{eq:D_time}
\mathbf{D}^{\text{Time}}(\tau)
=\mathrm{diag}\!\big(
\mathbf{R}(\tau\theta_1),\,
\mathbf{R}(\tau\theta_2),\,
\dots,\,
\mathbf{R}(\tau\theta_{d'/2})
\big),
\end{equation}
where $\theta_k=b^{-2(k-1)/(d'/2)}$ denotes the frequency for each block and $\mathbf{R}(\alpha)$ is the standard $2\times2$ rotation matrix ${\small(\cos\alpha,-\sin\alpha;\,\sin\alpha,\cos\alpha)}$.

Given a query $Q_i$ and key $K_j$ associated with timestamps $\tau_i$ and $\tau_j$, their time-modulated forms are:
\begin{equation}
\begin{aligned}
Q_i^{\text{Time}} &= \big(\mathbf{D}^{\text{Time}}(\tau_i)\big)^{\!\top} Q_i,\\
K_j^{\text{Time}} &= \big(\mathbf{D}^{\text{Time}}(\tau_j)\big)^{\!\top} K_j.
\end{aligned}
\end{equation}
The time-aware attention logit is then calculated as follows:
\begin{equation}
\begin{aligned}
Q_i^{\text{Time}} (K_j^{\text{Time}})^\top
&= Q_i^\top \mathbf{D}^{\text{Time}}(\tau_i)\,\mathbf{D}^{\text{Time}}(\tau_j)^{\!\top} K_j\\
&= Q_i^\top \mathbf{D}^{\text{Time}}(\tau_i - \tau_j)\, K_j,
\end{aligned}
\end{equation}
which explicitly encodes the continuous temporal offset. Aggregating tokens gives the time-aware attention equation:
\begin{equation}
\label{eq:attn_time_final}
\mathrm{Attn}^{\text{Time}}(Q,K,V)
=\mathrm{softmax}\!\left(
\frac{Q^{\text{Time}} (K^{\text{Time}})^{\!\top}}{\sqrt{d}}
\right)V.
\end{equation}
This continuous-time variant of RoPE injects world-time awareness directly into attention mechanism while introducing no additional learnable parameters.

\paragraph{Time-AdaLN.}
Because a DiT operates on patchified video tokens whose temporal resolution is downsampled relative to the original frame sequence, Time-RoPE alone cannot capture fine-grained frame-level timing. To preserve precise temporal control, we introduce an additional learnable pathway that injects the continuous time signal at the feature level. A 1D convolution $f_{\text{time}}(\cdot)$ encodes the frame-level timestamps $\tau_i$ into embeddings that are then injected through an Adaptive Layer Normalization (AdaLN) module via a learned affine transformation:
\begin{equation}
\tilde{z}'_{i,n} = \mathrm{LN}\!\big(\tilde{z}_{i,n}\big)\,\odot\, f_\gamma\!\big(f_{\text{time}}(\tau_i)\big)\;+\; f_\beta\!\big(f_{\text{time}}(\tau_i)\big),
\label{eqn:adaln_time}
\end{equation}
where $f_\gamma(\cdot)$ and $f_\beta(\cdot)$ are lightweight MLPs producing scale and shift, and $\odot$ denotes element-wise multiplication.

AdaLN is well-suited for temporal modulation since the world-time variable is a smooth global scalar that influences the dynamics of the entire scene rather than localized spatial regions. Compared to other conditioning mechanisms commonly used in video diffusion models, AdaLN achieves more stable and coherent temporal dynamics than cross-attention~\cite{esser2024scaling} and additive schemes~\cite{he2025cameractrl2,bai2025recammaster}, because it modulates activations with a single smooth global time signal rather than injecting token-level perturbations that can disrupt spatial alignment or produce unstable temporal responses (see ablations in \cref{sec:experiments:ablation}).

Combined with Time-RoPE, it forms our temporal backbone, providing both feature-wise and attention-level time control for stable and coherent video generation.

\subsection{4D Control}
\label{subsec:method:camera_control}
To enable full 4D controllability within the diffusion process, we extend the world-time–conditioned model with explicit camera viewpoint control through a unified 4D positional encoding (4D-RoPE) and a dedicated camera-conditioned normalization branch (Camera-AdaLN).

\paragraph{4D-RoPE.}
To jointly control time and camera viewpoint within attention, we extend Time-RoPE into a 4D positional encoding by incorporating a camera-aware geometric component. Following~\cite{li2025prope}, we encode relative camera-ray or camera-pose relationships to construct a camera-aware rotary transformation, which is combined with the temporal rotation to form a unified 4D operator. This 4D extension injects both continuous time differences and viewpoint-dependent geometric relations directly into the attention mechanism, enabling disentangled and coherent control of temporal evolution and camera motion during generation.

\paragraph{Camera-AdaLN.}
We introduce a parallel AdaLN branch for camera pose modulation. Following existing camera-controlled video diffusion models~\cite{he2025cameractrl2}, we encode per-pixel camera geometry using Pl\"{u}cker ray~\cite{plucker1865xvii} embeddings and aggregate them into token-level features via a 2D convolutional camera encoder. These features predict affine parameters through lightweight MLPs that modulate intermediate activations according to the desired camera trajectory.

\subsection{Dataset for Disentangled 4D Control}
\label{subsec:method:dataset}
Learning to control temporal dynamics and camera motion independently requires data that explicitly decouples variations along these two dimensions. However, existing video datasets are not designed for such disentangled learning. Single-camera datasets~\cite{bahmani2025ac3d} contain only one fixed or slowly moving viewpoint per sequence, inherently coupling temporal and spatial dimensions, while multi-camera datasets~\cite{bai2025recammaster} typically employ synchronized captures with uniform temporal sampling, where all cameras observe identical dynamics at the same timestamps.

To enable disentangled supervision, we construct a dataset that contains diverse temporal and spatial variations within each scene. For each scene, we generate multiple temporal variants by applying time-remapping functions (e.g., slow motion, pausing, random speed change) to moving objects, and render each temporal variant under different camera trajectories. We build the dataset using the PointOdyssey framework~\cite{zheng2023pointodyssey} within Blender, enabling physically consistent dynamic scenes with controllable cameras, lighting, and character motions. All videos are annotated with corresponding camera parameters and world-time labels for conditional training.
Figure~\ref{fig:overview} includes an example from our 4D-disentangled training dataset.

In total, the dataset comprises approximately 2k scenes across 80 environments and 100 characters, each with 3 camera trajectories and 3 temporal patterns, resulting in about 20k videos for learning controllable 4D video diffusion models. Please refer to the Appendix for dataset details and visualizations.

\begin{figure}[t]
	\centering
	\includegraphics[width=1.0\linewidth]{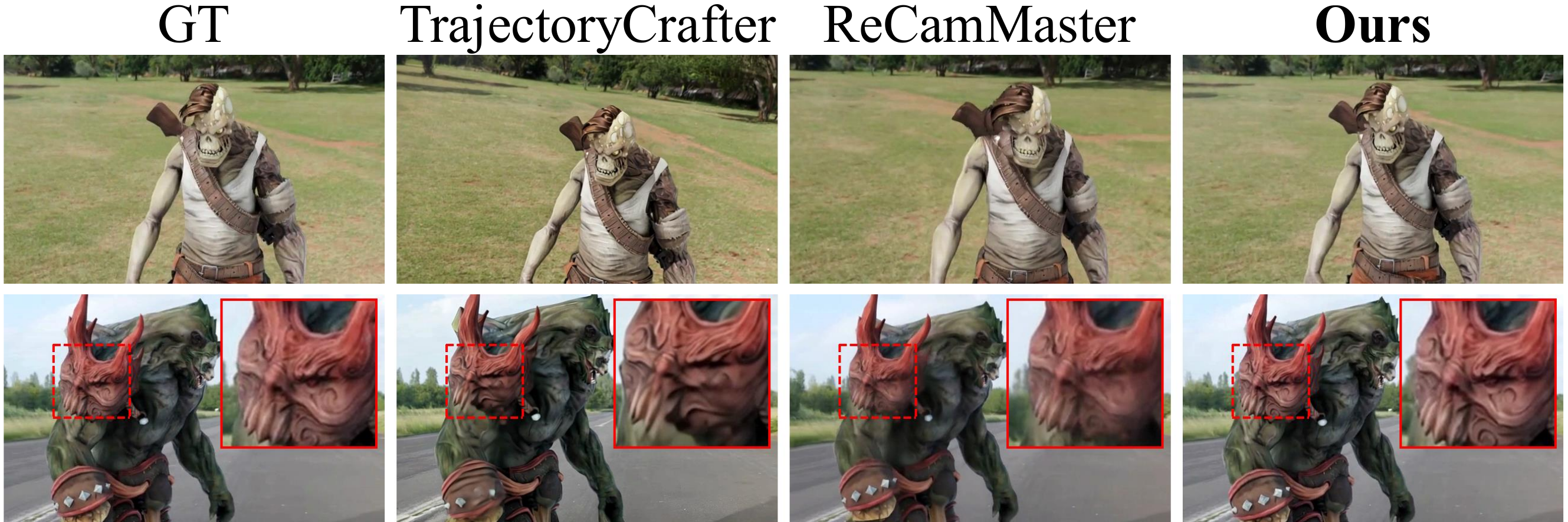}
	\vspace{-0.7cm}
    \caption{\textbf{Comparison on Synthetic Videos.} 
    GT frames compared with predictions from our method and state-of-the-art novel-view synthesis models. Our method adheres most closely to the target camera conditions and produces the finest level of detail.}
	\label{fig:time_slice_synthetic}
\end{figure}

\begin{table}[t]
\centering
\setlength{\tabcolsep}{2pt}
\caption{
    \textbf{Comparison of Camera- and Time-Controlled Video Generation on the Synthetic Dataset.} Baseline methods designed solely for camera control (denoted with *) are extended to 4D control by applying time remapping~\cite{huang2022rife} to the input videos prior to camera-conditioned generation. Our approach attains the highest pixel-level accuracy across all metrics, demonstrating its effectiveness in jointly modeling camera and temporal control.
}
\vspace{-0.2cm}
\small
\begin{tabular}{lccc}
\toprule
Method & PSNR$\uparrow$ & SSIM$\uparrow$ & LPIPS$\downarrow$ \\
\midrule
TrajectoryCrafter\textsuperscript{*} & 17.72 & 0.4917 & 0.3431 \\
ReCamMaster\textsuperscript{*}       & 21.86 & 0.5852 & 0.1846 \\
\rowcolor{cyan!10} Ours              & \textbf{24.57} & \textbf{0.6905} & \textbf{0.1265} \\
\bottomrule
\end{tabular}
\label{tab:cam_and_time_synthetic}
\end{table}

\begin{table*}[t]
\centering
\setlength{\tabcolsep}{2.0pt}
\caption{
    \textbf{Comparison of Camera- and Time-Controlled Video Generation on Real-World Videos.} Our method achieves the most accurate camera pose control and produces videos with reduced temporal flicker, smoother motion, and higher subject–background consistency, indicating stronger 4D controllability while maintaining high visual quality.
}
\vspace{-0.3cm}
\resizebox{\textwidth}{!}{
\begin{tabular}{lcccccccccc}
\toprule
 & \multicolumn{2}{c}{Camera Accuracy} & \multicolumn{6}{c}{VBench} & \multicolumn{2}{c}{Video Quality} \\
\cmidrule{2-3} \cmidrule(lr){4-9} \cmidrule{10-11}
\makecell[c]{Method} &
\makecell[c]{RotErr$\downarrow$} &
\makecell[c]{TransErr$\downarrow$} &
\makecell[c]{Aesthetic\\Quality$\uparrow$} &
\makecell[c]{Imaging\\Quality$\uparrow$} &
\makecell[c]{Temporal\\Flickering$\uparrow$} &
\makecell[c]{Motion\\Smoothness$\uparrow$} &
\makecell[c]{Subject\\Consistency$\uparrow$} &
\makecell[c]{Background\\Consistency$\uparrow$} &
\makecell[c]{FVD$\downarrow$} &
\makecell[c]{KVD$\downarrow$} \\
\midrule
TrajectoryCrafter*  & 5.44 & 3.31 & \textbf{0.4525} & \textbf{0.5696} & 0.9659 & 0.9881 & 0.9328 & 0.9375 & 2399 & 150.2 \\
ReCamMaster*        & 2.98 & 1.85 & 0.4470 & 0.5256 & 0.9755 & 0.9911 & 0.9375 & 0.9472 & 2325 & 146.1 \\
\rowcolor{cyan!10}
Ours                & \textbf{1.47} & \textbf{1.32} & 0.4520 & 0.5598 & \textbf{0.9780} & \textbf{0.9923} & \textbf{0.9428} & \textbf{0.9506} & \textbf{2292} & \textbf{139.1} \\
\bottomrule
\end{tabular}
}
\label{tab:cam_and_time_realworld}
\end{table*}

\begin{figure*}[t]
\centering
\includegraphics[width=1.0\linewidth]{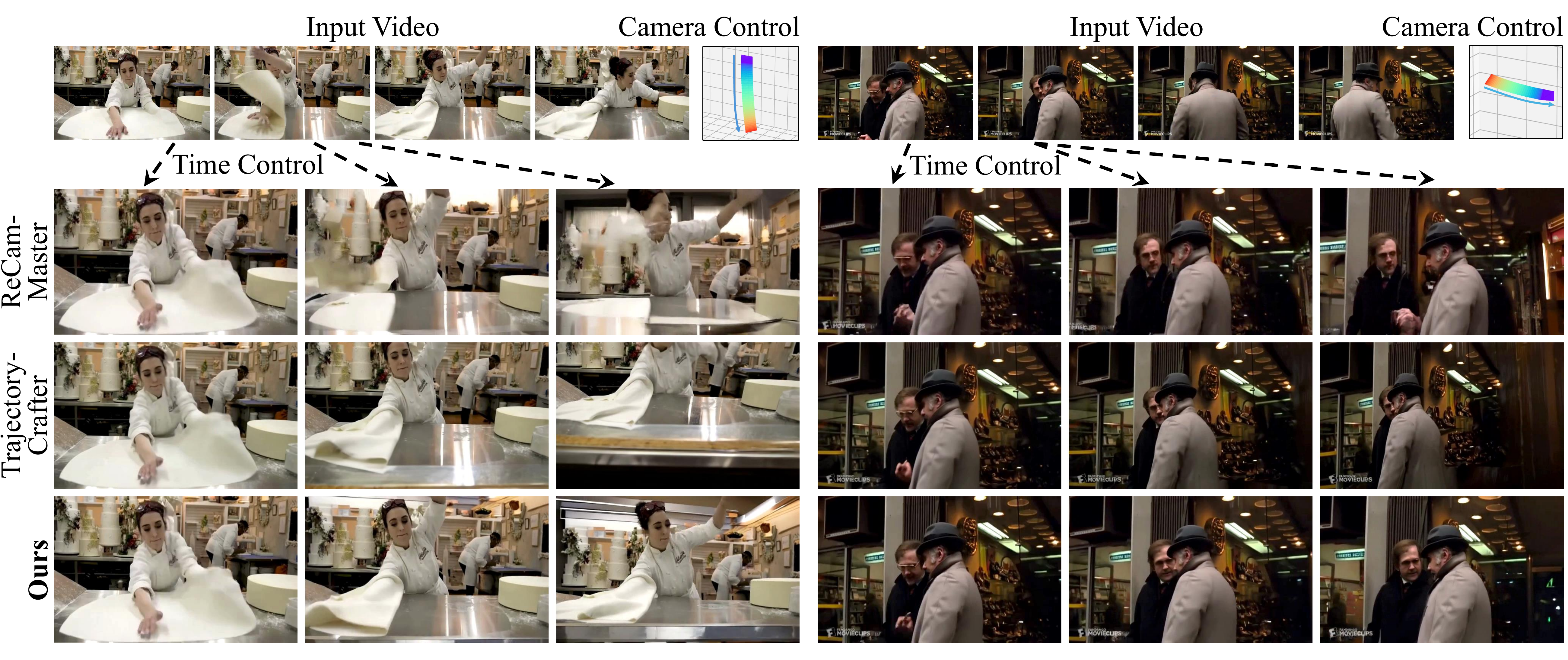}
\vspace{-0.85cm}
\caption{
    \textbf{Qualitative Comparison of Camera- and Time-Controlled Video Generation on Real-World Videos.} Qualitative comparison between our method and state-of-the-art novel-view synthesis models extended with time remapping~\cite{huang2022rife}. In the left example, existing methods struggle under extreme view and time changes, producing severe artifacts (ReCamMaster) and showing imprecise camera control (TrajectoryCrafter). The right example similarly illustrates strong artifacts and reduced detail from ReCamMaster, while TrajectoryCrafter again fails to follow the prescribed trajectory.
}
\label{fig:cmp_realworld}
\end{figure*}

\section{Experiments}
\label{sec:experiments}
Additional details and video results are provided in the supplementary material and accompanying video. Our dataset, code, and models will be publicly released.

\paragraph{Implementation Details.}
We build our model on top of CogVideoX~\cite{yang2024cogvideox}, using the pretrained CogVideoX-5B-T2V as the base model.
For video-to-video generation conditioned on an input video, we follow ReCamMaster~\cite{bai2025recammaster} and concatenate the source and target video tokens along the frame dimension. We then fine-tune the base model with our 4D camera–time control module.
All experiments are conducted at a spatial resolution of $384 \times 640$ with 81 frames per video.
We adopt a progressive training schedule that first trains the model at half resolution and subsequently fine-tunes it at full resolution.
The final model is trained for a total of 40K iterations with a batch size of 64.

\paragraph{Metrics and Baselines.}
We evaluate 4D-controlled video-to-video generation on both synthetic and real-world videos.
For synthetic evaluation, we generate 500 videos using PointOdyssey~\cite{zheng2023pointodyssey}, featuring unseen characters and unseen scenes. For real-world evaluation, we collect 100 videos from ViPE~\cite{huang2025vipe}.
For the synthetic dataset, we report PSNR, SSIM, and LPIPS to assess  reconstruction quality under joint camera and time control.
For each real-world video, we randomly sample unseen camera trajectories and temporal patterns to generate the outputs.
We evaluate visual quality using VBench~\cite{huang2024vbench} metrics as well as FVD and KVD~\cite{ge2022long}. 
We follow the evaluation protocol of previous camera-controlled video generation methods~\cite{bai2025recammaster, he2024cameractrl, he2025cameractrl2} by measuring rotation and translation errors. Specifically, we compare the ground-truth camera trajectories against the camera poses estimated using MegaSAM~\cite{li2025megasam} from generated videos.

For comparison, we extend state-of-the-art camera-controlled video-to-video generation models~\cite{bai2025recammaster, yu2025trajectorycrafter} to support joint time and camera control. Specifically, we apply time remapping~\cite{huang2022rife} to the input video to produce the target temporal pattern, and then perform camera-conditioned generation using each baseline. We mark this adaption of the baselines by adding the * symbol.

\begin{figure*}[t]
\centering
\includegraphics[width=1.0\linewidth]{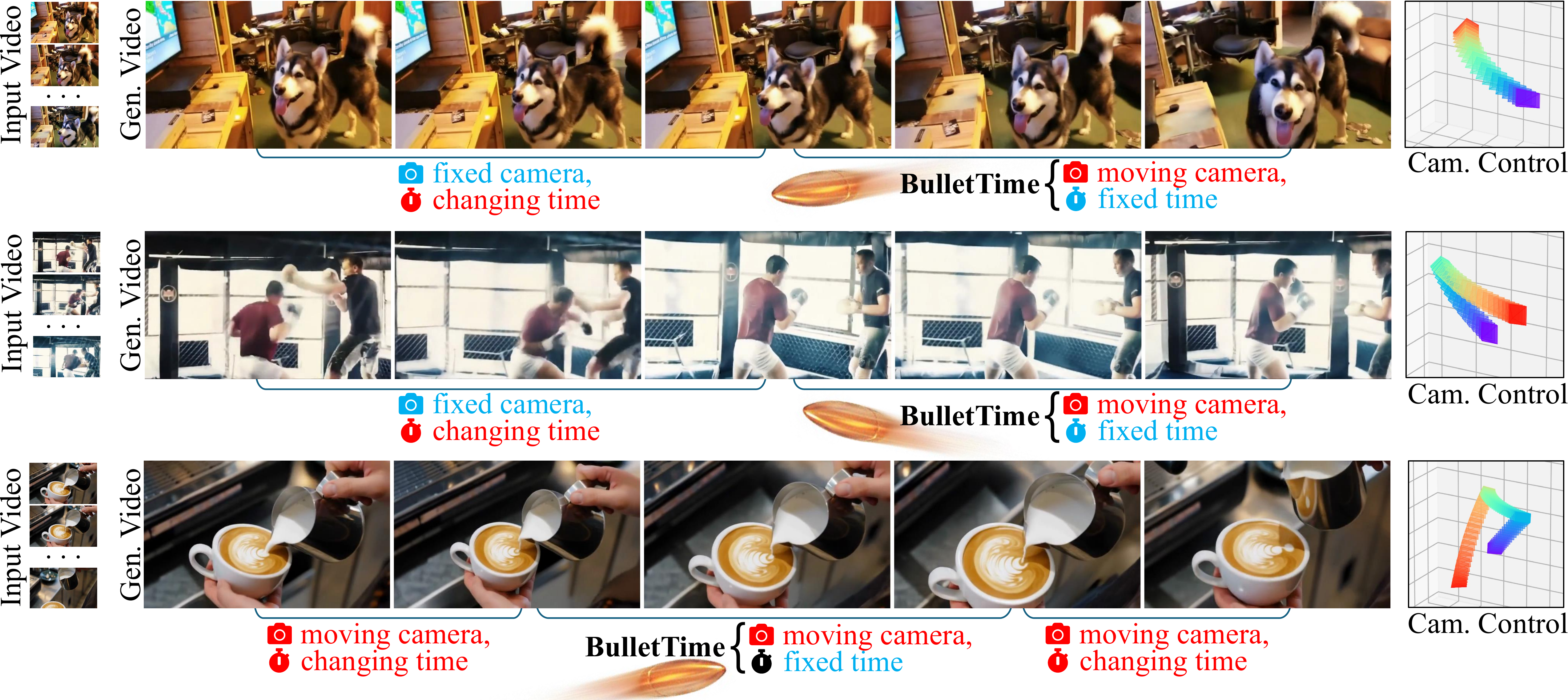}
\vspace{-0.7cm}
\caption{
    \textbf{4D Control: Camera and Time Manipulation.} Our model generates videos that faithfully follow independently specified camera and time controls. Each row shows combinations of fixed or moving camera viewpoints (\faCamera) and fixed or changing world time (\faStopwatch). The model correctly applies each control mode, including challenging settings such as moving camera with fixed time (\textbf{bullet time} effect), while preserving scene dynamics and visual coherence. These results indicate strong disentanglement between camera and world time conditioning as well as robust generalization across diverse real world inputs.
}
\label{fig:diverse_cam_scenes}
\end{figure*}

\subsection{Evaluation}
We conduct a comprehensive evaluation of our model’s ability to perform 4D-controlled video generation, assessing how well it generates videos that faithfully follow explicitly specified camera and time controls given an input sequence.

\paragraph{Comparison on Synthetic Videos.}
We first evaluate all methods on the synthetic dataset, where ground-truth videos provide a direct reference for assessing 4D-controlled generation. For a fair comparison, we fine-tune ReCamMaster on our 4D-controlled dataset, yielding substantially improved performance over its released checkpoint (e.g., more than a 2 dB gain in PSNR; see Appendix for details). As shown in \cref{tab:cam_and_time_synthetic}, our method achieves the highest pixel-level accuracy across all metrics, reflecting its ability to jointly model camera and time control. Qualitative results in \cref{fig:time_slice_synthetic} further show that our approach follows the prescribed camera paths and temporal inputs more faithfully while producing sharper and more detailed generations than the baselines.

\paragraph{Comparison on Real-World Videos.}
We further evaluate our method on real-world videos, assessing both camera-pose consistency and overall visual quality following the evaluation protocols of prior work~\cite{bai2025recammaster, he2025cameractrl2}. As shown in \cref{tab:cam_and_time_realworld}, our approach achieves the lowest rotation and translation errors, indicating more accurate camera control when jointly conditioning on time. It also attains higher temporal stability and better subject–background consistency according to VBench~\cite{huang2024vbench} while maintaining overall visual quality comparable to existing approaches. Although TrajectoryCrafter reports slightly higher image-quality scores on VBench, we find that its camera control is substantially less reliable. For reference, the input conditional videos obtain FVD=2012 and KVD=131.8 on UCF101~\cite{soomro2012ucf101}.

Qualitative comparisons in \cref{fig:cmp_realworld} corroborate these findings. All models are asked to generate videos under jointly specified camera and time controls. TrajectoryCrafter’s reliance on monocular-depth-based point-cloud projections often results in geometric distortions and inaccurate camera trajectories. Conversely, ReCamMaster~\cite{bai2025recammaster} exhibits noticeable artifacts in dynamic regions once time control is introduced. Our method, in contrast, produces stable scene dynamics and faithfully follows both the prescribed temporal pattern and camera motion.

\begin{figure}[t]
\centering
\includegraphics[width=1.0\linewidth]{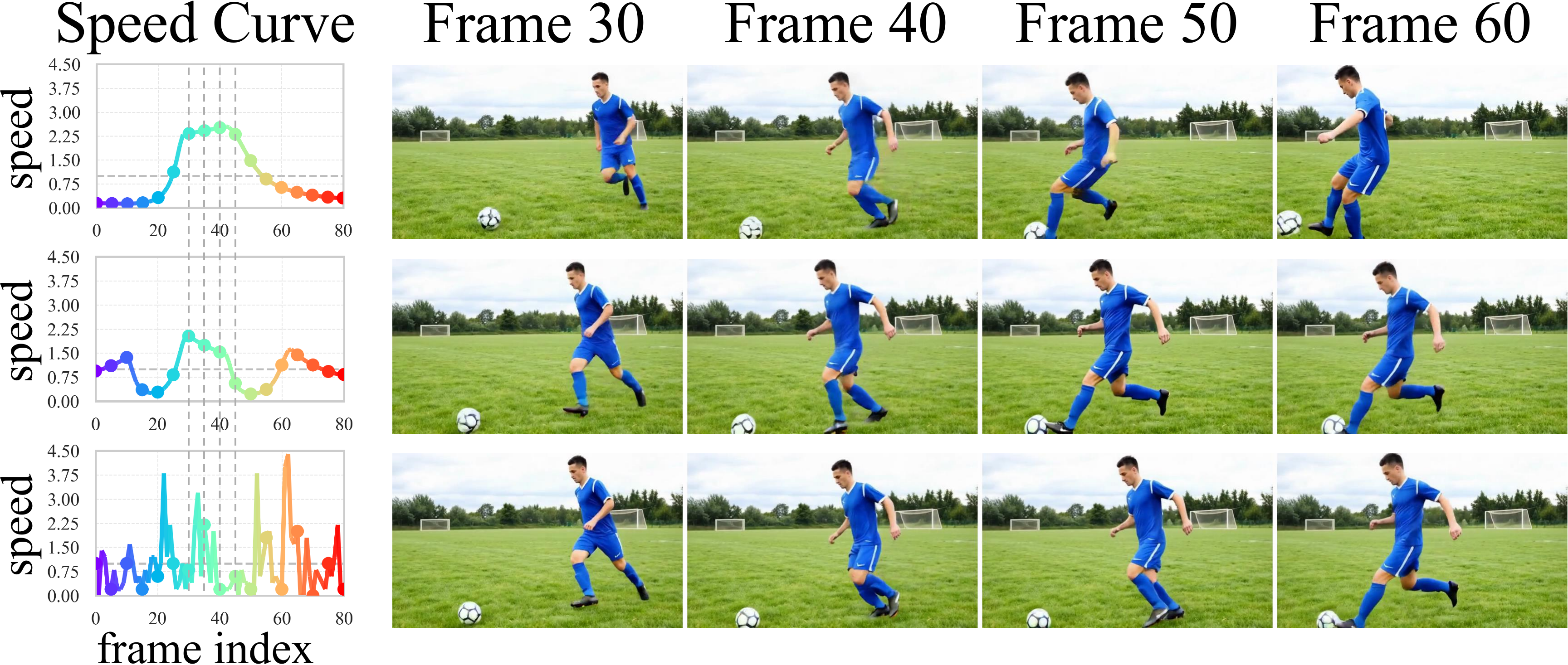}
\vspace{-0.7cm}
    \caption{\textbf{Time Control Generalization.} Three generations produced by our method from the same input video under different time conditions. Although the model is trained on only a limited subset of time remappings, it generalizes well to complex and previously unseen temporal inputs.
}
\label{fig:diverse_time_control}
\end{figure}

\subsection{Discussion}

\paragraph{Advantages over Camera-Only Methods.}
To extend camera-only video-to-video methods~\cite{bai2025recammaster, yu2025trajectorycrafter} to joint time–camera control, prior approaches rely on time remapping the input video before performing camera-conditioned generation. However, this two-stage pipeline causes the conditioning video itself to change across different time-control settings, leading to inconsistent 4D generation and, in some cases, loss of input content (e.g., when applying bullet-time effects, later parts of the video are clipped due to temporal pausing). 
In contrast, our end-to-end 4D-controllable video diffusion model conditions on the same input video for all camera and time settings, ensuring stable supervision and consistent content preservation.
To quantify this advantage, we measure background consistency on synthetic videos under identical camera controls but varying time inputs. Foreground masks are obtained via SAM2~\cite{ravi2024sam2} and ground-truth first-frame mattes, and metrics are computed over background regions to assess visual similarity. As shown in \cref{tab:cam_and_time_disentanglement}, our method achieves substantially higher scores, reflecting stronger camera–time disentanglement and improved content consistency. Qualitative results in \cref{fig:cam_and_time_disentanglement} further illustrate the superior 4D consistency our model achieves on real-world videos.

\paragraph{Generalization across Scenes, Cameras, and Timings.}
We also demonstrate that our method generalizes effectively to diverse real-world scenarios involving unseen camera trajectories and temporal variations, despite being fine-tuned exclusively on a synthetic human-centric dataset. As shown in \cref{fig:diverse_cam_scenes}, our approach robustly handles a wide range of camera inputs and maintains coherent 4D control across varied environments, extending beyond human subjects to animals and scenes with complex physical dynamics. Moreover, \cref{fig:diverse_time_control} illustrates that our model accommodates a broad set of previously unseen temporal control patterns, further highlighting its strong generalization capability.

\subsection{Ablation Study}
\label{sec:experiments:ablation}

\paragraph{World-time Conditoning.}
We study alternative mechanisms for conditioning continuous world time and compare them with our proposed Time-RoPE and AdaLN. 
We conduct time-controlled video-to-video generation experiments on the synthetic dataset of~\cite{bai2024syncammaster}, where camera poses are fixed and only time conditioning is applied.

We compare AdaLN with two commonly used learnable conditioning strategies in video diffusion models: (1) \emph{Cross-Attention}, where temporal conditions are treated as additional tokens that interact with visual tokens through a cross-attention layer, following multimodal diffusion architectures such as MMDiT~\cite{esser2024scaling}; and (2) \emph{Channel Addition}, where temporal embeddings are directly added to the latent feature channels, following prior camera-controlled video generation methods~\cite{he2025cameractrl2,bai2025recammaster}. We also compare our Time-RoPE with the original RoPE to assess the benefit of injecting continuous world time into attention.   

\begin{table}[t]
\centering
\setlength{\tabcolsep}{2pt}
\caption{
\textbf{Evaluation of Disentangled Camera and Time Control.}
Background consistency is measured between videos generated under identical camera trajectories but different time controls using masked image metrics~\cite{gao2022dycheck} (mPSNR, mMAE, mSSIM, mLPIPS), where the prefix “m” indicates evaluation within the masked background region.
Our method achieves higher consistency than ReCamMaster, indicating more effective disentanglement and improved 4D coherence.
}
\vspace{-0.2cm}
\small
\begin{tabular}{lcccc}
\toprule
Method & mMAE$\downarrow$ & mPSNR$\uparrow$ & mSSIM$\uparrow$ & mLPIPS$\downarrow$ \\
\midrule
ReCamMaster\textsuperscript{*} & 0.0362 & 25.80 & 0.8789 & 0.1527 \\
\rowcolor{cyan!10} Ours & \textbf{0.0231} & \textbf{28.29} & \textbf{0.9096} & \textbf{0.1119} \\
\bottomrule
\end{tabular}
\label{tab:cam_and_time_disentanglement}
\end{table}
\begin{figure}[t]
\centering
\vspace{0.4cm}
\includegraphics[width=1.0\linewidth]{Figures/results/bg_consistency.pdf} 
\vspace{-0.65cm}
\caption{
    \textbf{Comparison of Camera–Time Disentanglement.} When varying the time condition while keeping the camera condition fixed, state-of-the-art camera-controlled video generation methods such as ReCamMaster fail to maintain consistent camera control, resulting in geometric inconsistencies within the generated content.
}
\label{fig:cam_and_time_disentanglement}
\end{figure}

As shown in ~\cref{tab:ablation_time} (rows 1–3 and 5–7), AdaLN delivers the strongest temporal control among the learnable conditioning baselines, both with the original RoPE and with Time-RoPE. Time-RoPE on its own (row 4) already provides strong temporal control and surpasses all learnable variants that use standard RoPE (rows 1–3). Because Time-RoPE injects a time-control prior directly into the attention logits, replacing RoPE with Time-RoPE consistently improves performance across all learnable conditioning methods. The combination of Time-RoPE with AdaLN yields the best results on all metrics, highlighting the effectiveness of our proposed world-time conditioning design.
%

\paragraph{4D Conditioning.}
We analyze the effect of our 4D-RoPE and camera/time AdaLN in adding 4D conditioning. \cref{tab:ablation_cam_and_time} shows that removing either module degrades performance, with a more substantial drop observed when omitting 4D-RoPE. All models are trained with a batch size of 4 for 20k iterations, using the same dataset configuration as \cref{tab:cam_and_time_synthetic}.

\begin{table}[t]
\centering
\setlength{\tabcolsep}{2pt}
\caption{
    \textbf{Ablation on World-Time Conditioning.} We compare different temporal conditioning mechanisms for fine-tuning CogVideoX~\cite{yang2024cogvideox} toward world-time-controlled video-to-video generation. AdaLN provides the strongest performance among the learnable baselines (CrossAttention and ChannelAddition). Time-RoPE itself offers strong controllability and consistently improves upon standard RoPE across all configurations, with Time-RoPE + AdaLN achieving the best overall results.
}
\vspace{-0.2cm}
\small
\begin{tabular}{lccc}
\toprule
Method & PSNR$\uparrow$ & SSIM$\uparrow$ & LPIPS$\downarrow$ \\
\midrule
RoPE + CrossAttention       & 23.86  & 0.8274 & 0.1753 \\
RoPE + ChannelAddition        & 25.31  & 0.8438 & 0.1456 \\
RoPE + AdaLN                & 29.83  & 0.8821 & 0.0742 \\
\midrule
Time-RoPE                   & 30.45  & 0.8807 & 0.0753 \\
Time-RoPE + CrossAttention  & 30.51  & 0.8816 & 0.0753 \\
Time-RoPE + ChannelAddition   & 30.40  & 0.8813 & 0.0730 \\
\rowcolor{cyan!10} Time-RoPE + AdaLN & \textbf{32.15} & \textbf{0.8962} & \textbf{0.0631} \\
\bottomrule
\end{tabular}
\label{tab:ablation_time}
\end{table}

\begin{table}[t]
\setlength{\tabcolsep}{6pt}
\vspace{0.35cm}
\centering
\small
\caption{\textbf{Ablation on 4D Conditioning.}
Our 4D-RoPE and Time/Camera AdaLN enhance 4D conditioning scores. 
}
\label{tab:ablation_cam_and_time}
\vspace{-0.25cm}
\begin{tabular}{lccc}
\toprule
Method & PSNR$\uparrow$ & SSIM$\uparrow$ & LPIPS$\downarrow$ \\
\midrule
\rowcolor{cyan!10} Full Model & \textbf{23.45} & \textbf{0.6283} & \textbf{0.1309}     \\
\quad w/o 4D-RoPE                  & 21.98    & 0.6099 & 0.1785 \\
\quad w/o Camera/Time AdaLN        & 22.74    & 0.6197 & 0.1493 \\
\bottomrule
\end{tabular}
\end{table}


\section{Conclusion}
We presented a 4D-controllable video diffusion framework that disentangles scene dynamics from camera motion through continuous world-time conditioning and explicit viewpoint control. Our framework injects continuous world-time and camera-trajectory signals via a unified 4D positional encoding and adaptive layer normalization modules, which together provide effective conditioning as confirmed by our ablation studies. Supported by a synthetic dataset designed to vary temporal and camera factors independently, our approach enables precise manipulation of scene dynamics and flexible camera navigation within dynamic environments. Experimental results demonstrate that this disentangled control achieves state-of-the-art performance in 4D-controllable video generation while remaining fully compatible with existing video diffusion architectures. 

Despite these advances, several challenges remain. Our model relies on a synthetic dataset for disentangled supervision, which, although effective, may not capture the full complexity of real-world physics, lighting, and long-horizon scene dynamics with large camera baseline. In addition, while our method supports arbitrary time and camera inputs, inference is still performed in a parallel (non-autoregressive) diffusion setting, limiting its ability to model extremely long videos or persistent worlds. Future work could explore autoregressive or recurrent formulations of 4D-controllable diffusion, enabling temporally unbounded generation and online trajectory control. Another promising direction is learning disentanglement together with real-world video corpora. Extending the framework to handle physics-aware temporal reasoning may further broaden its applicability.

We hope this work inspires continued research toward scalable, controllable, and physically consistent 4D generative video models.

\section*{Acknowledgements}
We gratefully acknowledge the insightful discussions with Qin Han, Yutong Chen, and Haofei Xu.
This work was supported as part of
the Swiss AI initiative by a grant from the Swiss National
Supercomputing Centre (CSCS) under project IDs \#36 on Alps, enabling large-scale training.
We thank the NVIDIA Academic Grant Program for providing GPU resources for this project.

{
    \small
    \bibliographystyle{ieeenat_fullname}
    \bibliography{main}
}

\clearpage

\maketitlesupplementary
\appendix
\setcounter{page}{1}
\setcounter{table}{0}
\setcounter{figure}{0}
\setcounter{equation}{0}
\counterwithin{figure}{section}
\counterwithin{table}{section}
\renewcommand{\thetable}{\thesection.\arabic{table}}
\renewcommand{\thefigure}{\thesection.\arabic{figure}}
\renewcommand{\theequation}{\thesection.\arabic{equation}}

\section{Additional Implementation Details}
We fine-tune the pretrained CogVideoX-5B-T2V~\cite{yang2024cogvideox} model using the AdamW optimizer with a learning rate of \(2\times10^{-5}\) and a weight decay of \(10^{-4}\). 
We apply gradient clipping with a maximum norm of \(1.0\). 
The learning rate follows a linear decay schedule with 100 warm-up steps. 
The model is trained for a total of 40K iterations.

Since existing latent-space video diffusion models downsample the temporal resolution via the 3D-VAE and the transformer’s patchify block, directly injecting per-frame world-time signals becomes incompatible with the reduced temporal resolution. To address this, we encode the world-time sequence with learnable 1D convolution layers to map it to the downsampled temporal resolution.
For our non-learnable Time-RoPE design, we downsample the world-time sequence via average pooling so that it matches the latent temporal resolution. When computing the RoPE positional encoding, we scale the world-time differences by the FPS to convert them into consistent temporal offsets. With this scaling, Time-RoPE becomes exactly equivalent to the standard RoPE formulation under uniform time sampling, making standard RoPE a special case of our design to better preserves the model’s original generative prior.
For camera-conditioning signals, we follow prior work~\cite{bai2025recammaster, he2025cameractrl2} and directly subsample the camera trajectories at fixed intervals to match the latent temporal resolution.

For video distribution-level evaluation, we report FVD and KVD following the protocol of TATS~\cite{ge2022long}. We use videos from UCF101~\cite{soomro2012ucf101} as the real distribution and generate camera–time–conditioned outputs for each baseline as the fake distribution. Following TATS, each video is embedded using a pretrained I3D network to obtain a single feature vector, and we compute FVD (Fréchet distance) and KVD (polynomial-kernel MMD) between the real and generated feature distributions. Using UCF101 as the reference distribution introduces several limitations: many of our time-control operations (e.g., slow motion, temporal pausing) fall outside the temporal statistics of UCF101, potentially inflating divergences unrelated to perceptual quality; and UCF101 primarily contains human-action videos with limited camera motion, whereas our generated videos may include large, user-specified camera trajectories. 

\section{Additional Results}
We provide full video results in the accompanying supplementary video and recommend watching it for additional qualitative demonstrations.

\begin{figure}[htp]
	\centering
	\includegraphics[width=1.0\linewidth]{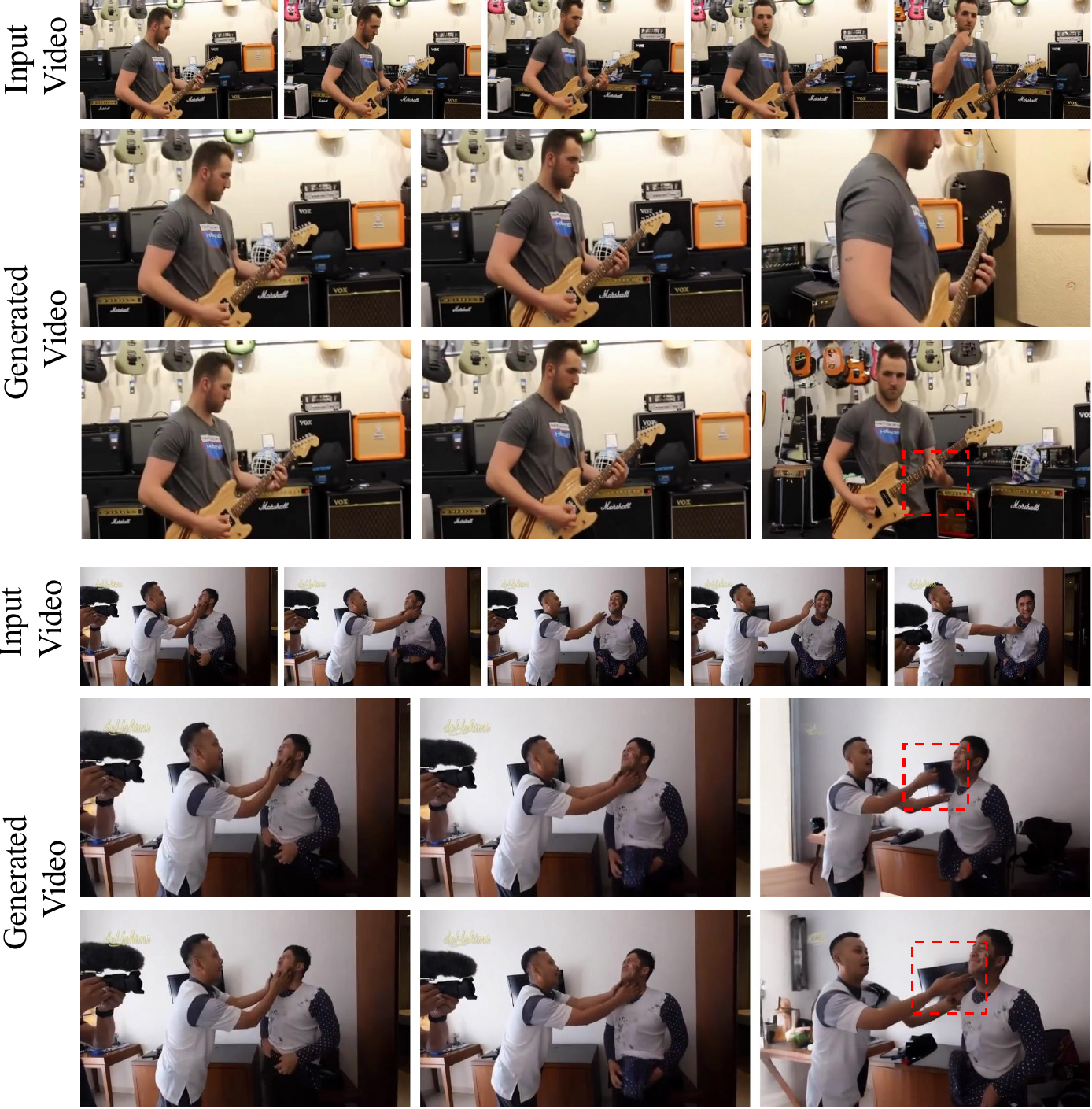}
    \caption{\textbf{Failure Cases.} Our model may struggle with fine-grained hand motion and with generating high-quality background details under certain viewpoints.}
	\label{fig:failure_case}
\end{figure}

\paragraph{Failure Cases}
Since our model is fine-tuned from a pre-trained video diffusion model, it inherits some of their generation limitations. For example, it may struggle with fine-grained hand details under certain viewpoints, where motions may violate physical plausibility or appear low-quality. 
In addition, background regions that are not visible in the input video may lack high-fidelity details because our training data is limited to synthetic environments. This limitation could potentially be alleviated by jointly incorporating real-world datasets during training.
\cref{fig:failure_case} shows examples of these failures. 

\section{4D Controlled Dataset}

We have curated our 4D-controlled dataset using PointOdyssey~\cite{zheng2023pointodyssey} and its released 3D assets, including character models, HDR environment textures, and indoor scene layouts.
Specifically, we use 100 human-like characters and animate them using real-world motion capture data. 
The characters are paired with 3D environments that include both outdoor and indoor scenes. For outdoor scenes, we use 60 HDR environment textures to simulate natural backgrounds, while for indoor scenes, we use 20 unique 3D environments provided by PointOdyssey.

\begin{figure*}[ht]
\centering
\includegraphics[width=1.0\linewidth]{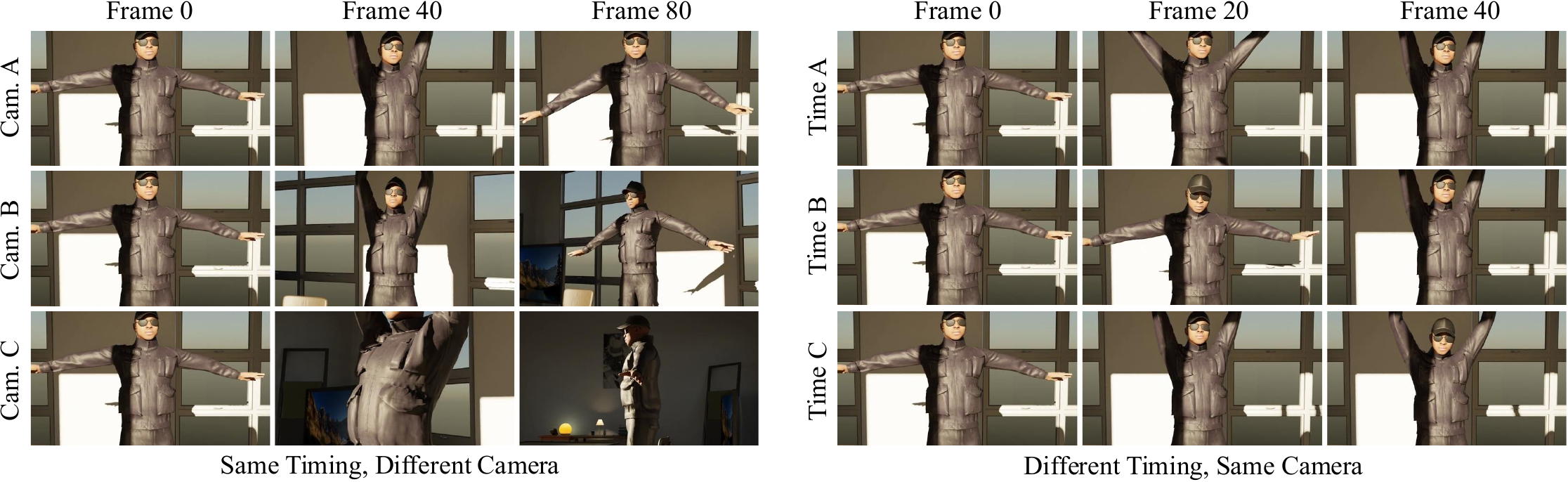}
\vspace{-0.7cm}
\caption{
\textbf{4D Controlled Data Sample}. We demonstrate the decoupled curation of spatial control (camera motion) and temporal control (timing). (\textit{Left}) Three distinct camera views (Cam. A, B, C) show the same synchronized frames (Frame 0, 40, 80), confirming simultaneous camera and action control. (\textit{Right}) Three videos with different action timings (Time A, B, C) are generated from the same camera trajectory, showing independent control over the temporal dimension. 
}
\label{fig:4dcontrolled_dataset}
\end{figure*}

\begin{figure*}[ht]
\centering
\includegraphics[width=1.0\linewidth]{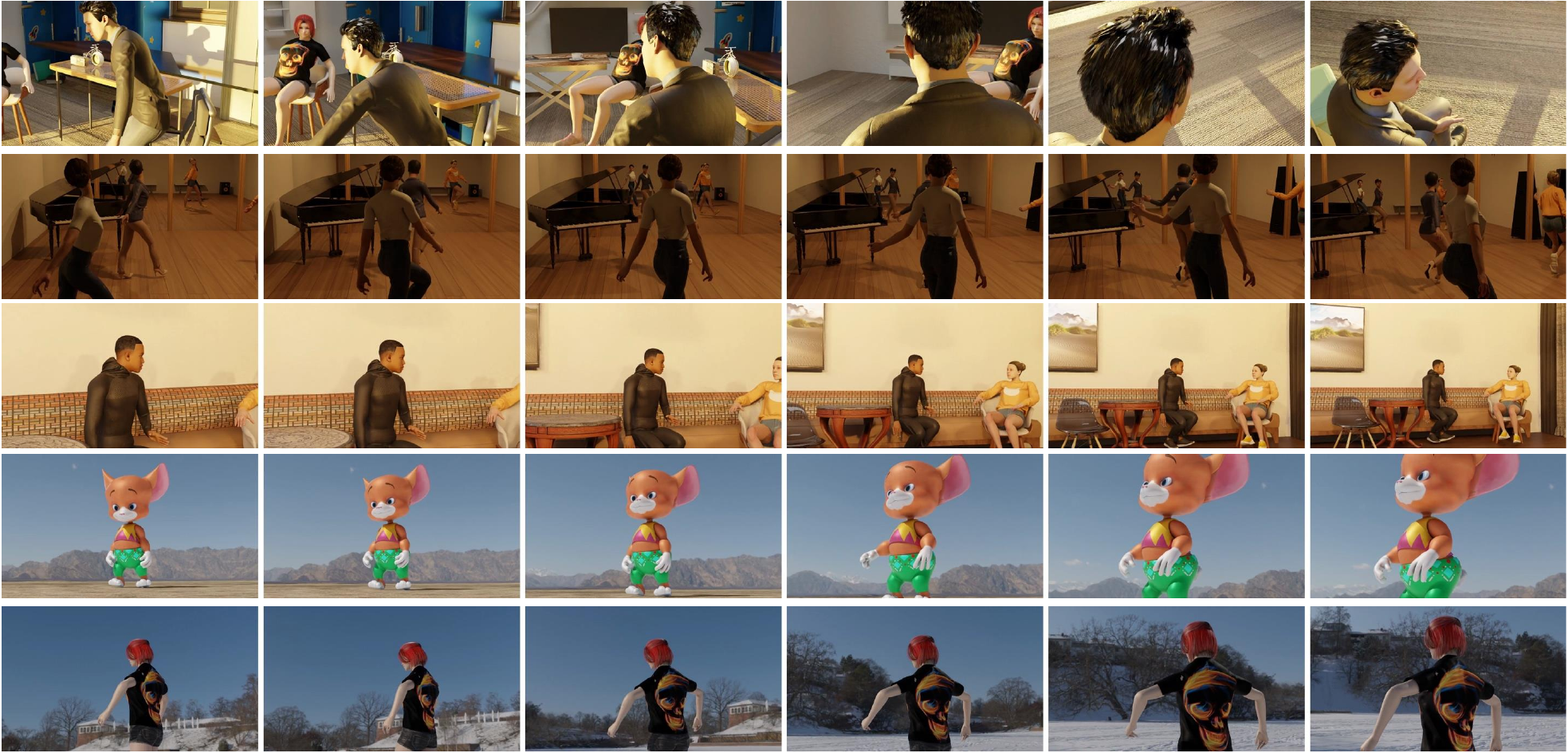}
\vspace{-0.7cm}
\caption{
\textbf{Examples of Our 4D Controlled Data Samples.} We showcase the diversity of our generated dataset, which spans diverse motions, a variety of human-like subjects, both single- and multi-character scenes, and a wide range of indoor and outdoor environments. We will release the dataset and its generation code.
}
\label{fig:dataset_samples}
\vspace{-0.2cm}
\end{figure*}

\begingroup
\renewcommand{\thefigure}{D.1}
\begin{figure*}[ht!]
\centering
\includegraphics[width=1.0\linewidth]{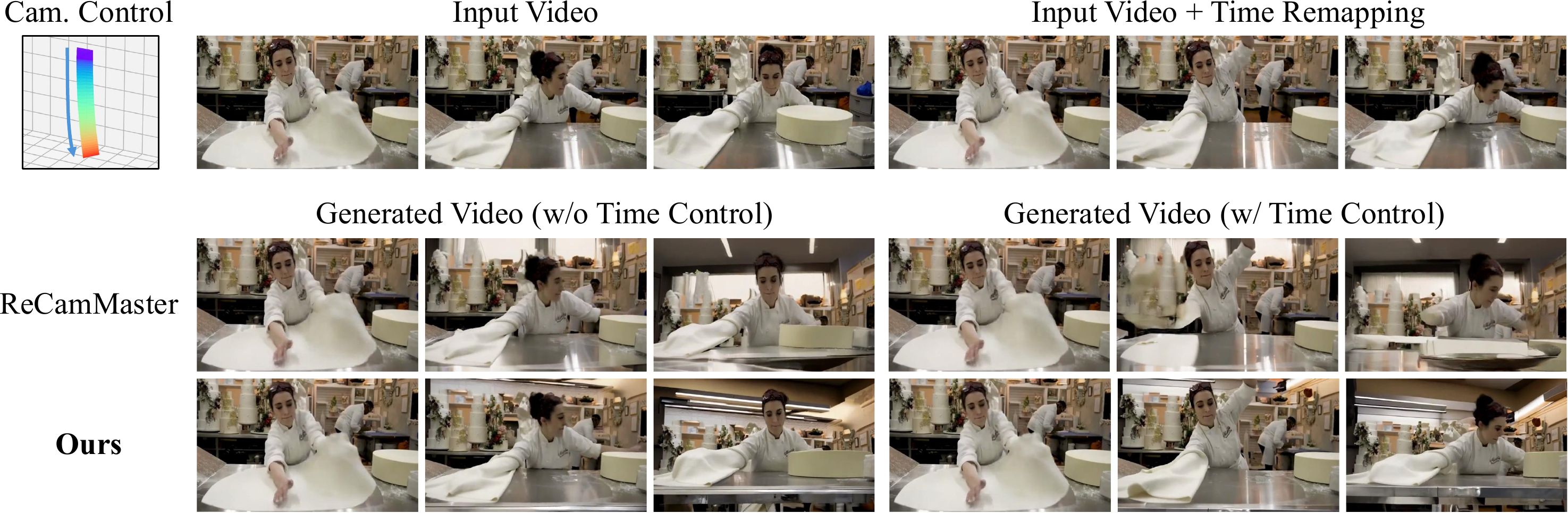}
\vspace{-0.4cm}
\caption{\textbf{ReCamMaster Artifacts under Time Remapping.}
In the left three columns we show how ReCamMaster and our method perform under no time remapping. The right three columns show the same generated scene with time remapping. We can observe how ReCamMaster creates significant artifacts after applying time remapping. In comapison our method does not suffer from this.}
\vspace{-0.2cm}
\label{fig:recam_no_time}
\end{figure*}

\addtocounter{figure}{1}
\endgroup

For each character–scene pair, we introduce independent control over camera viewpoints and world-time to generate 4D-controlled sequences.
We first define linear world-time as the original motion-capture timeline, where the motion progresses at a constant rate. To enable world-time control, we replace this linear progression with a remapped timeline that changes how the animation advances over time. The character animation is then generated by retargeting the mocap sequence according to this remapped timeline. By accelerating, decelerating, or holding specific portions of the timeline, these remappings allow effects such as slow motion, pausing, and locally accelerated motion segments.
Speicfically, %
To realize these behaviors, we construct several temporal variants that introduce diverse world-time dynamics. These include a slow-motion variant, where consecutive output frames map to closely spaced points on the mocap timeline, effectively reducing the motion speed, and a pausing variant, where selected poses are held for extended durations.
In addition, we use a random time-warping variant that maps the linear timeline to an upsampled one while enforcing local speed constraints, and a spline-based variant, where monotonic spline control points are sampled under slope constraints to produce smooth, continuously varying changes in motion speed.
Together, these variants provide diverse and natural temporal augmentations and enable the model to learn a wide range of world-time controls. For each scene, we curate data using one linear timeline and two additional variants randomly selected from the above set.

We determine each camera position using a look-at center, a radius, and a pair of rotation angles representing azimuth and elevation. A static camera keeps these parameters fixed throughout the sequence. For dynamic trajectories, we sample 2--4 waypoint cameras by perturbing the look-at center, radius, and rotation angles, and then interpolate these waypoints to obtain a smooth camera path. To ensure realistic framing, we enforce specific constraints across all trajectories: the radius is sampled within a range of $4$--$12$ meters, the total azimuth variation is limited to $75^\circ$, the elevation variation to $30^\circ$, and the look-at center offset to a maximum of $1$ meter from the human character's centroid. For each scene, we generate three distinct views: one trajectories with more than two waypoints, one orbit-style trajectory with a single waypoint, and one static camera. To produce natural, eased motion between waypoints, we randomly select either uniform-speed interpolation or a smoothstep function, defined as $f(t) = 3t^{2} - 2t^{3}$ for $t \in [0,1]$. We use fixed camera intrinsic settings with a focal length of $30\,\mathrm{mm}$ and a sensor width of $50\,\mathrm{mm}$.

\section{Baselines}

\paragraph{ReCamMaster~\cite{bai2025recammaster}.}
We utilize the official codebase and the released checkpoint, which was pre-trained on 122k synthetic videos. To ensure a fair comparison on our synthetic benchmark, we fine-tune ReCamMaster on our dataset of 20k synthetic videos. As demonstrated in \cref{tab:finetune_recam}, while fine-tuning yields substantial performance gains, the method still significantly underperforms compared to our approach. This highlights the superior effectiveness of our proposed 4D control module.

We further observe that visual artifacts in ReCamMaster results stem primarily from the attempt to enforce time-control effects, as illustrated in \cref{fig:recam_no_time}. When time control is disabled (top row), ReCamMaster exhibits significantly fewer artifacts compared to scenarios where time control is active (bottom row). Crucially, these artifacts persist even after fine-tuning on our time-controlled patterns. Given that time manipulation is a prerequisite for full 4D controllability, these results suggest that ReCamMaster's two-stage pipeline suffers from intrinsic limitations in handling complex temporal dynamics.


\begin{table}[htpb]
\centering
\setlength{\tabcolsep}{4pt}
\caption{
\textbf{Finetuning ReCamMaster on Our Dataset. }
}
\vspace{-0.25cm}
\begin{tabular}{lcccc}
\toprule
 & PSNR$\uparrow$ & SSIM$\uparrow$ & LPIPS$\downarrow$ \\
\midrule
ReCamMaster  & 19.67 & 0.5426 & 0.2594 \\
 \hspace{1em} + finetune on our data  & 21.86 & 0.5852 & 0.1846  \\
\rowcolor{cyan!10} Ours              & \textbf{24.57} & \textbf{0.6905} & \textbf{0.1265} \\ 
\bottomrule
\end{tabular}
\label{tab:finetune_recam}
\end{table}

\paragraph{TrajectoryCrafter~\cite{yu2025trajectorycrafter}.} 
We employ the official codebase and the provided checkpoint, trained on approximately 180k videos. TrajectoryCrafter relies on the reconstruction of dynamic point clouds from monocular videos. However, inaccuracies in these point clouds—often stemming from unreliable depth estimation—frequently lead to geometric artifacts, a limitation also noted in the original paper. Our experimental results show similar artifact cases driven by depth inconsistencies.

\end{document}